# Identifying factors associated with fast visual field progression in patients with ocular hypertension based on unsupervised machine learning

Xiaoqin Huang[1], PhD, Asma Poursoroush[1], MSc, Jian Sun[2], PhD, Michael V. Boland[3], MD, PhD, Chris Johnson[4], PhD, and Siamak Yousefi[1,5], PhD

[1]Department of Ophthalmology, University of Tennessee Health Science Center, Memphis, USA
[2]Integrated Data Sciences Section, Research Technologies Branch, National Institute of Allergy and Infectious Diseases, National Institutes of Health (NIH), Bethesda, USA
[3]Department of Ophthalmology, Massachusetts Eye and Ear, Boston, USA
[4]Department of Ophthalmology and Visual Sciences, University of Iowa Hospitals and Clinics, Iowa City, IA
[5]Department of Genetics, Genomics, and Informatics, University of Tennessee Health Science Center, Memphis, USA

**Corresponding Author:**
Siamak Yousefi, 930 Madison Ave., Suite 726, Memphis, TN 38163
Phone: 9014487831, Email: siamak.yousefi@uthsc.edu

## Abstract

**Purpose:** To identify ocular hypertension (OHT) subtypes with different trends of visual field (VF) progression based on unsupervised machine learning and to discover factors associated with fast VF progression.

**Design:** Cross-sectional and longitudinal study.

**Participants**: A total of 3133 eyes of 1568 ocular hypertension treatment study (OHTS) participants with at least five follow-up VF tests were included in the study.

**Methods:** We used a latent class mixed model (LCMM) to identify OHT subtypes using standard automated perimetry (SAP) mean deviation (MD) trajectories. We characterized the subtypes based on demographic, clinical, ocular, and VF factors at the baseline. We then identified factors driving fast VF progression using generalized estimating equation (GEE) and justified findings qualitatively and quantitatively.

**Main Outcome Measure:** Rates of SAP mean deviation (MD) change.

**Results:** The LCMM model discovered four clusters (subtypes) of eyes with different trajectories of MD worsening. The number of eyes in clusters were 794 (25%), 1675 (54%), 531 (17%) and 133 (4%). We labeled the clusters as Improvers, Stables, Slow progressors, and Fast progressors based on their mean of MD decline, which were 0.08, -0.06, -0.21, and -0.45 dB/year, respectively. Eyes with fast VF progression had higher baseline age, intraocular pressure (IOP), pattern standard deviation (PSD) and refractive error (RE), but lower central corneal thickness (CCT). Fast progression was associated with calcium channel blockers, being male, heart disease history, diabetes history, African American race, stroke history, and migraine headaches.

**Conclusion:** Unsupervised clustering can objectively identify OHT subtypes including those with fast VF worsening without human expert intervention. Fast VF progression was associated with higher history of stroke, heart disease, diabetes, and history of more using calcium channel blockers. Fast progressors were more from African American race and males and had higher incidence of glaucoma conversion. Subtyping can provide guidance for adjusting treatment plans to slow vision loss and improve quality of life of patients with a faster progression course.

**Introduction**

Glaucoma is a heterogeneous group of disorders that represents the second leading cause of blindness worldwide.[1] Glaucoma will become more prevalent in the US in the coming decades due to the aging population. Currently, glaucoma-related expenses are estimated to be over $1 billion annually and there is a sharp increase in the costs of late-stage glaucoma treatment[2,3]. Therefore, developing reliable methods for identifying those at higher risk of glaucoma progression and future vision loss is critical.

Progression rates based on visual field (VF) tests have been used as functional surrogate endpoints in most recent glaucoma clinical trials including the ocular hypertension treatment study (OHTS).[4,5] Identifying patients who progress fast is paramount to implement (more aggressive) treatment regimens to slow future vision loss.

Most of the current studies have focused on determining progression of each individual subject based on ordinary least square (OLS) estimation of VF test points, global parameters, or patterns of VF loss (derived via unsupervised machine learning models) rates. [6-16] These models use all patients' data (as a single group) to estimate thresholds to identify progressing and non-progressing groups. However, rather than using information from a single population data, it may be beneficial to segregate subgroups of patients with similar VF progressing trends then assigning patients to different subtypes (clusters) based on their characteristics to tailor monitoring and treatment.

To this end, mixed models include a fixed-effect component that represents the characteristics of a population and a random-effect component that reflects the degree of deviation of an individual eye from the mean population. This process creates eye-specific intercepts and slopes. Although not yet incorporated in routine clinical practice, estimates of rates of change using linear mixed models have been widely applied in clinical research[3–8].

Several statistical methods have been developed to identify subgroups of trajectories of a quantitative marker over time including latent class linear mixed model (LCMM). [17] The LCMM is an extension of the standard linear mixed model for handling various subtypes of longitudinal trajectories. LCMM captures the heterogeneity in individual trajectories and identifies subgroups of samples with similar profiles of trends, independently of observed sample's characteristics.[18] LCMM has been successfully used in identification of subtypes in renal function trajectories[19], identifying longitudinal-growth patterns from infancy to childhood[20], and identifying subgroups of disease trajectories with differential patterns of genetic association[21]. The LCMM has been recently used to predict perimetric loss in patients with glaucoma.[22] This study aims to leverage unsupervised LCMM to identify different subtypes of OHT patients based on their VF loss trajectories and to identify the baseline factors that may drive fast VF progression.

## Methods

### Dataset

The study was performed in accordance with the tenets of the Declaration of Helsinki. Appropriate data use agreements were signed to use the deidentified data from the OHTS. The OHTS was a multicenter randomized trial designed to evaluate the safety and efficacy of topical ocular hypertensive medication in delaying or preventing the onset of primary open-angle glaucoma (POAG) in individuals with elevated intraocular pressure (IOP) and no detectable glaucomatous damage [23].

The OHTS recruited 1636 participants with OHT from 22 sites. Each participant was examined about twice a year using Humphrey standard automated perimetry (SAP) 30-2 VF testing. The demographic and clinical characteristics included age, self-reported race and ethnicity, self-reported sex, IOP, central corneal thickness (CCT), and refractive error (RE)[24]. We included 3133 eyes from 1568 subjects with at least five reliable longitudinal VFs for the downstream analysis. As we had access to only the first two phases of the OHTS, we included longitudinal data from year 1994 to 2009.

### Latent-class mixed model (LCMM)

As an extension of linear mixed model, LCMM includes two joint sub-models: a multinomial logistic regression model and a linear mixed model. The multinomial logistic regression model expresses the probability for each eye belonging to each cluster[25,26] and the linear mixed model is specific to each latent class where it expressed the observed mean deviation (MD) of each eye at specific time point as the sum of the expected MD value and the individual eye's departure (represented by random effects) from that expected MD value. The LCMM assumes that the heterogeneous trajectories are composed of $G$ clusters of subjects characterized by $G$ mean profiles of trajectories. Each subject (trajectory) belongs to only one cluster so the cluster membership is defined by a discrete random variable $c_i$ that equals $g$ if subject $i$ belongs to the cluster $g$ ($g = 1, \ldots, G$).

$$Y_{ij}|_{c_i=g} = X_{L_1i}(t_{ij})^T \beta + X_{L_2i}(t_{ij})^T V_g + Z_i(t_{ij})^T u_{ig} + W_i(t_{ij}) + e_{ij}$$

$X_{Li}(t_{ij})$ and $Z_i(t_{ij})$ are two vectors of covariates at time $t_{ij}$. $X_{L1i}(t_{ij})$ is the vector associated with common fixed effects $\beta$ over clusters and $X_{L2i}(t_{ij})$ is the vector associated with cluster-specific fixed effects $v_g$. The vector $Z_i(t_{ij})$, which typically includes functions of time $t_{ij}$, is associated with the vector of random effects $u_i$. Finally, the process $(w_i(t))_{t \in R}$ is a zero mean Gaussian stochastic process, $e_{ij}$ is the measurement errors[18]. LCMM considers both linear and non-linear dynamics of the MD trend by using the link function (we selected "beta" as the link function based on preliminary results).

### Identification of subtypes (clusters) and validation

We applied LCMM on the trajectory of the SAP MD of eyes and identified various clusters. We then calculated the posterior probability of an eye belonging to each cluster and generated subtypes. We evaluated the robustness of the entire clustering model based on different approaches. For identifying the optimal number of clusters, we evaluated the Integrated Completed Likelihood (ICL) metric and selected the optimal number of clusters corresponding to the minimum ICL. To assure the selected number of clusters is consistent, we evaluated other metrics including Akaike Information Criterion (AIC) and Bayesian Information Criterion (BIC) as well. To investigate cluster repeatability, we randomly selected subsets of the eyes and performed clustering again. We then compared the clustering memberships with the initial clustering membership and assessed repeatability. Regarding the number of visits, we excluded several visits (last 1-3 visits) from trajectories, and re-applied clustering and evaluated the membership consistency same as above. LCMM models were implemented in R using the "lcmm" package [18].

**Characterizing subtypes**

To label clusters, we computed the mean MD rates of eyes in each cluster, and labeled clusters based on mean rate of MD change. To characterize clusters, we compared different parameters of the eyes in each cluster. More specifically, we investigated age, race, gender, IOP, CCT and other ocular and systemic parameters (at the baseline visit) as well as medications of eyes in clusters. We used generalized estimating equation (GEE) to compare continuous and categorical variables among clusters accounting for both eyes of same subjects at patient level. We also computed the odds ratio of the categorical variables in each cluster. The statistical analysis was performed in SPSS 28.0 (IBM Corp, Armonk, NY).

**Results**

**Demographic characteristics**

The mean age of participants at the baseline was 56.0 years (SD: 9.5). The average baseline MD and IOP of all eyes were 0.01 dB (SD: 1.4) and 24.9 mmHg (SD: 3.0), respectively. The average number of visits was 22.3 (SD: 6.52), with an average follow-up time of 11.59 (SD: 3.07) years. Table 1 shows the detailed characteristics of the patients at the baseline visit.

**Table 1** Demographic characteristics of the participants (eyes) at the baseline visit. Continuous variables are summarized based on mean and standard deviation (SD) while categorical variables are presented as counts and frequencies (%).

| Variable | Value |
|---|---|
| Age (Years) | 56.0 (9.5) |
| Intraocular pressure (IOP; mmHg) | 24.9 (3.0) |
| Central corneal thickness (CCT; Micron) | 572.8 (38.9) |
| Mean deviation (MD; dB) | 0.01 (1.4) |
| Pattern standard deviation (PSD; dB) | 2.0 (0.49) |
| Cup-to-disc ration (CDR) | 0.36 (0.19) |
| Vertical CDR | 0.39 (0.20) |
| Refractive Error (RE, Diopter) | -0.66 (2.4) |
| Cases converted to POAG (1) | 361 (11.5%) |
| POAG conversion based on VF (1) | 204 (6.5%) |
| POAG conversion based on optic disc (1) | 287 (9.5%) |
| POAG conversion based on VF and od (1) | 130 (4.2%) |
| Prescribed medication (1) | 768 (24.5%) |
| High blood pressure history (1) | 1179 (37.6%) |
| Prescribed TCAI (1) | 17 (0.54%) |
| Grandparent has glaucoma (1) | 116 (7.9%) |
| Aunt or uncle has glaucoma (1) | 388 (12.4%) |
| Diabetes history (1) | 367 (11.7%) |
| Prescribed Alpha (1) | 1 (0.03%) |
| Other conditions history (1) | 885 (28.3%) |
| Cancer history (1) | 177 (5.7%) |
| Sibling has glaucoma (1) | 335 (10.7%) |
| Prescribed miotics (1) | 8 (0.26%) |
| Any family history glaucoma (1) | 1315 (42.0%) |
| Prescribed Epinephrin/dipiverefin (1) | 19 (0.61%) |
| Systemic beta blockers (1) | 140 (4.5%) |
| Parent/sibling family history glaucoma (1) | 1059 (33.8%) |
| Low blood pressure history (1) | 138 (4.4%) |
| African American or Other race (1) | 767 (24.5%) |
| Asthma history (1) | 218 (7.00%) |
| Education (b'5') | 664 (21.2%) |
| Heart disease history (1) | 192 (6.1%) |
| Active medication (1) | 1547 (49.4%) |
| Male (GENDER-1) | 1350 (43.1%) |
| Parents have glaucoma (1) | 881 (28.1%) |
| Calcium channel blockers (1) | 369 (11.8%) |
| Stroke history (1) | 32 (1.0%) |
| Chronic lung disease (1) | 72 (2.3%) |
| Migraine headaches (1) | 344 (11.0%) |

**Clusters identified based on LCMM**

The LCMM model discovered four subtypes (clusters) of eyes with different courses of VF loss. The number of eyes in clusters 1 to 4 were 794 (25%), 1675 (54%), 531 (17%), and 133 (4%), respectively. The probability of sample membership was consistently higher than 0.80 for eyes in all clusters. The mean slope of MD of eyes in clusters 1 to 4 were 0.08 dB/year, -0.06 dB/year, -0.21 dB/year and -0.45 dB/year, respectively (Table 2). According to their MD rate of change, clusters 1 to 4 were labeled as Improvers, Stables, Slow-progressors, and Fast progressors, respectively (Fig.1). Randomly selecting 40% to 90% of the eyes and performing clustering led to over 93.8% correct memberships reflecting a high membership accuracy and robust clustering (Table 3).

**Table 2.** Intercept and slope of the model for each class.

| Cluster | Intercept (CI) | Slope (CI) |
|---|---|---|
| 1 | 0 (-0.04, 0.03) | 0.083 (0.073, 0.092) |
| 2 | 0.68 (0.64, 0.71) | -0.060 (-0.071, -0.048) |
| 3 | 0.93 (0.91, 0.95) | -0.21 (-0.23, -0.20) |
| 4 | 0.98 (0.95, 1.00) | -0.45 (-0.47, -0.42) |

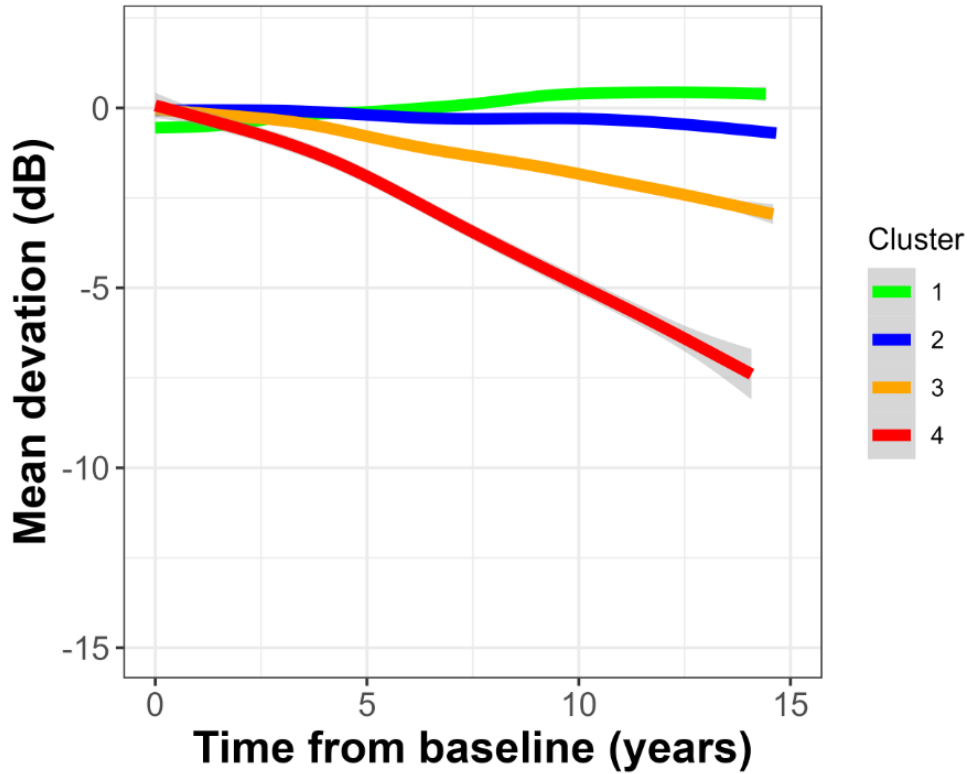

**Figure 1** Four clusters (subtypes) identified by the LCMM model. Smoothed trajectory of MD of the eyes in four clusters representing Improvers (green), Stables (blue), Slow progressors (brown), and Fast progressors (red).

**Table 3.** Membership accuracy result.

| Subsampling percentage (%) | Accuracy % (CI) |
|---|---|
| 90 | 95.6 (93.2, 97.6) |
| 80 | 94.0 (91.7, 96.3) |
| 70 | 94.5 (92.2, 96.8) |
| 60 | 97.7 (95.3, 99.7) |
| 50 | 97.4 (94.8, 99.4) |
| 40 | 93.8 (91.2, 96.8) |
| Number of last visits excluded | Accuracy (%) |
| 1 | 0.91 |
| 2 | 0.86 |
| 3 | 0.80 |

## Characteristics of the subtypes (clusters)

Table 4 shows the statistics of some of the parameters (continuous variables) at baseline in each cluster. Subjects in cluster 1 were the youngest and subjects in clusters 2-4 were older consistently with subjects in cluster 4 becoming the oldest group. The average baseline IOP of eyes in subtypes Improvers, Stables, Slow-progressors and Fast-progressors (clusters 1 to 4) were 24.7, 24.8, 25.4, and 25.7 mmHg, respectively. The mean CCT of eyes in clusters Improvers, Stables, Slow-progressors, and Fast progressors were 574.3, 574.1, 570.5, and 557.5 μm, respectively. Baseline MD was worse in the Improvers subtype when compared to the other three clusters. Baseline PSD was higher in the Fast-progressors subtype but similar in Improver, Stables and Slow-progressors subtypes. While eyes in the Improvers subtype had the worst RE, eyes in the Fast-progressors had the best RE. Cup-disc-ratio (CDR) and vertical CDR (VCDR) were similar among eyes in different clusters.

**Table 4** Characteristics of the baseline parameters of the eyes in each cluster presented as mean (SD).

| Variable | Cluster 1 | Cluster 2 | Cluster 3 | Cluster 4 | [a] P value | [b] P value (4 vs. 12) | [c] Odds ratio (Cluster 4) |
|---|---|---|---|---|---|---|---|
| Age (years) | 51.8 (8.3) | 55.8 (9.2) | 61.3 (8.7) | 63.9 (9.0) | **0.00** | 0.00 | **1.10 (1.07,1.12)** |
| IOP (mmHg) | 24.7 (2.8) | 24.8 (2.9) | 25.4 (3.2) | 25.7 (3.3) | **<0.001** | 0.006 | **1.09 (1.02,1.17)** |
| CCT (micron) | 574.3 (37.5) | 574.1 (38.5) | 570.5 (41.2) | 557.5 (38.7) | **0.01** | **<0.001** | **0.99 (0.98,0.99)** |
| PSD (dB) | 2.0 (0.58) | 2.0 (0.48) | 2.0 (0.37) | 2.2 (0.36) | 0.14 | **<0.001** | **1.55 (1.30,1.86)** |
| RE (Diopter) | -1.0 (2.3) | -0.68 (2.5) | -0.24 (2.3) | 0.09 (2.0) | **<0.001** | **<0.001** | **1.17 (1.07,1.28)** |
| MD (dB) | -0.45 (1.4) | 0.14 (1.4) | 0.3 (1.4) | 0.2 (1.4) | **<0.001** | 0.11 | 1.10 (0.95,1.27) |
| CDR | 0.36 (0.19) | 0.36 (0.19) | 0.37 (0.20) | 0.36 (0.19) | 0.46 | 0.86 | 1.07 (0.38,3.05) |
| VCDR | 0.38 (0.20) | 0.39 (0.20) | 0.39 (0.22) | 0.40 (0.21) | 0.21 | 0.48 | 1.07 (0.38, 3.05) |

IOP: Intraocular pressure; CCT: Central corneal thickness; MD: Mean deviation; PSD: Pattern standard deviation; CDR: Cup disk ratio; VCDR: Vertical cup disc ratio; RE: Refractive error. a: significance among 4 clusters; b: difference between cluster 4 and combined clusters 1 and 2; c: odds ratio of cluster 4.

In addition to comparing the significance of difference of various factors among all subtypes, we also compared factors that were significantly different between Fast-progressors (cluster 4) and non-progressors (combined clusters 1 and 2) subtypes.

Age, IOP, CCT, PSD and RE were significantly different between Fast-progressors and Non-progressors subtypes. Based on the Fast-progressors subtype, age, IOP, PSD and RE were positively associated with fast progression while CCT was negatively associated with fast progression. CCT was significantly different between fast-progressors and non-progressors when we accounted for age.

Figure 2 shows the survival analysis of the eyes in four subtypes based on Kaplan-Meier approach. As expected, eyes in the Improvers subtype had the lowest likelihood and eyes in the Fast-progressors subtype had the highest likelihood of eventually converting to glaucoma.

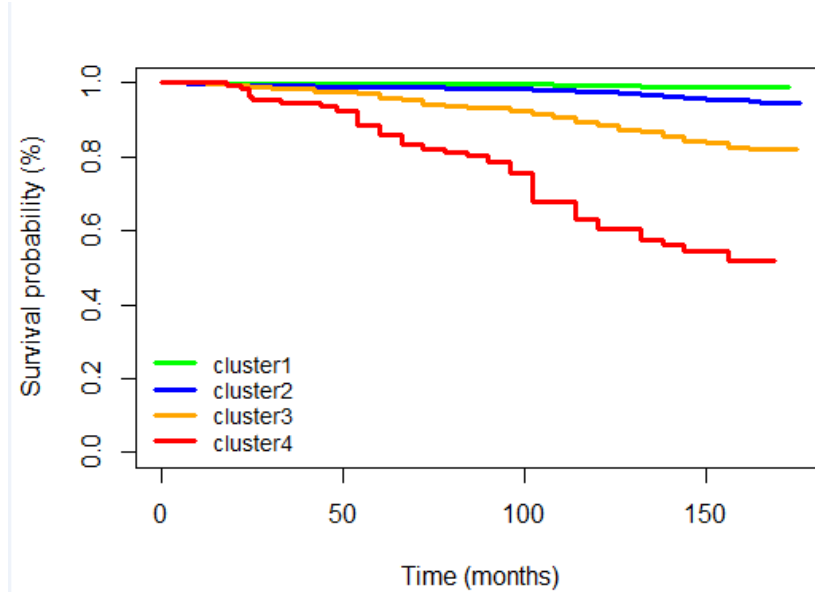

**Figure 2** Survival analysis of the eyes in different clusters.

Table 5 shows the characteristics of some of the categorical demographic, ocular, and systemic factors of subjects in different subtypes. We excluded the frequency of variables less than 1% in the statistical analysis, which included prescribed topical carbonic anhydrase inhibitors (TCAI), prescribed alpha, prescribed miotics, prescribed Epinephrine/ Dipivefrine). The percentage of eyes that eventually converted to glaucoma based on VF or optic disc abnormalities in Improvers, Stables, Slow-progressors, and Fast progressors subtypes (clusters 1 to 4) were 3.9%, 9.1%, 23.0% and 41.4%, respectively. Similar increasing trend of conversion rate was observed in subtypes once glaucoma was evaluated based on VF abnormality only, optic disc abnormality only, and both VF and optic disc abnormalities.

**Table 5.** Characteristics of eyes in different clusters---categorical variables.

| Variable | Cluster 1 (%) | Cluster 2 (%) | Cluster 3 (%) | Cluster 4 (%) | [a] P value | [b] P (4 vs 12) | [c] Odds ratio (Cluster 4) |
|---|---|---|---|---|---|---|---|
| POAG conversion (1) | 3.9 | 9.1 | 23.0 | 41.4 | **0.00** | 0 | 8.75 (5.89,13.01) |
| POAG conversion based on VF (1) | 1.0 | 3.8 | 15.6 | 36.8 | **0.00** | 0 | 19.41 (12.47, 30.22) |
| POAG conversion based on Disc (1) | 3.3 | 7.6 | 17.0 | 33.1 | **0.00** | 0 | 7.48 (4.88, 11.47) |
| POAG conversion based on VF and Disc (1) | 0.38 | 2.3 | 9.6 | 28.6 | **0.00** | 0 | 23.68 (14.03, 9.97) |
| Calcium channel blockers (1) | 7.4 | 12.8 | 12.6 | 21.8 | **<0.001** | 0.002 | 2.24 (1.35, 3.72) |
| Gender (male) | 40.4 | 42.1 | 46.5 | 57.9 | **0.01** | 0.002 | 1.93 (2.27, 2.94) |
| Heart disease history (1) | 4.0 | 6.2 | 7.5 | 12.8 | **0.00** | 0.004 | 2.55 (1.34, 4.86) |
| Diabetes history (1) | 7.6 | 11.4 | 17.0 | 20.3 | **<0.001** | 0.004 | 2.26 (1.29, 3.95) |

| Variable | | | | | | | |
|---|---|---|---|---|---|---|---|
| Other conditions history (1) | 25.2 | 28.2 | 30.6 | 39.4 | **0.02** | **0.011** | **1.74 (1.13, 2.67)** |
| African American race (1) | 28.6 | 22.2 | 23.2 | 34.6 | 0.26 | **0.025** | **1.65 (1.06, 2.57)** |
| Stroke history (1) | 0.50 | 1.2 | 0.56 | 3.8 | 0.17 | **0.036** | **3.98 (1.10, 14.41)** |
| Migraine headaches (1) | 11.0 | 12.0 | 9.4 | 4.5 | 0.24 | **0.04** | **0.36 (0.14, 0.94)** |
| Siblings have glaucoma (1) | 7.4 | 12.0 | 11.9 | 9.0 | **0.04** | 0.63 | 0.82 (0.38, 1.80) |
| High blood pressure history (1) | 32.2 | 38.7 | 41.6 | 41.4 | **0.00** | 0.365 | 1.22 (0.80, 1.86) |
| Cancer history (1) | 3.9 | 5.6 | 8.3 | 6.8 | **0.01** | 0.43 | 1.37 (0.63, 3.00) |
| Chronic lung disease history (1) | 2.0 | 2.4 | 2.6 | 1.5 | 0.70 | 0.56 | 0.66 (0.16, 2.73) |
| Aunt or uncle has glaucoma (1) | 10.8 | 13.2 | 11.3 | 15.8 | 0.57 | 0.42 | 1.26 (0.72, 2.20) |
| Asthma history (1) | 6.6 | 7.7 | 4.9 | 8.3 | 0.70 | 0.751 | 1.14 (0.52, 2.51) |
| Low blood pressure history (1) | 4.7 | 4.2 | 4.0 | 6.8 | 0.90 | 0.29 | 1.58 (0.68, 3.70) |
| Active medication (1) | 51.3 | 48.3 | 51.8 | 42.9 | 0.57 | 0.23 | 0.77 (0.51, 1.18) |
| Education (5- post graduate) | 23.1 | 20.1 | 22.0 | 20.3 | 0.48 | 0.85 | 0.95 (0.58, 1.57) |
| Any family history glaucoma (1) | 40.1 | 44.10 | 38.6 | 42.1 | 0.98 | 0.91 | 0.98 (0.64, 1.49) |
| Prescribed medication (1) | 25.2 | 23.8 | 26.6 | 21.1 | 0.94 | 0.37 | 0.83 (0.56, 1.25) |
| Grandparent has glaucoma (1) | 9.2 | 7.4 | 6.1 | 11.9 | 0.48 | 0.38 | 1.53 (0.60, 3.92) |
| Parent/sib family history glaucoma (1) | 31.9 | 35.8 | 31.6 | 28.6 | 0.90 | 0.24 | 0.76 (0.48, 1.21) |
| Parents have glaucoma (1) | 28.7 | 29.10 | 25.4 | 23.3 | 0.24 | 0.26 | 0.75 (0.46, 1.23) |
| Systemic beta blockers (1) | 3.4 | 4.6 | 4.9 | 7.5 | 0.12 | 0.13 | 1.85 (0.83, 4.13) |

Glaucoma conversion rate, calcium channel blockers, being male, heart disease history, diabetes history, other conditions history, siblings have glaucoma, high blood pressure history and cancer history were significantly different among four subtypes. Figure 3 shows the Chord diagram of clinical variables with odds ratio greater than 1 of eyes in the Fast-progressors subtype (cluster 4). When compared Fast-progressors subtype with Non-progressors subtype (combined clusters 1 and 2), glaucoma conversion ratio, calcium channel blockers, being male, heart disease history, diabetes history, stroke history, migraine headaches, other conditions history, and African American race were significantly different. The odds ratio of eyes in the Fast-progressors subtype implied that the glaucoma conversion ratio, calcium channel blockers, being male, heart disease history, diabetes history, stroke history, other conditions history, and African American race were positively associated while migraine headache was negatively associated with fast VF progression.

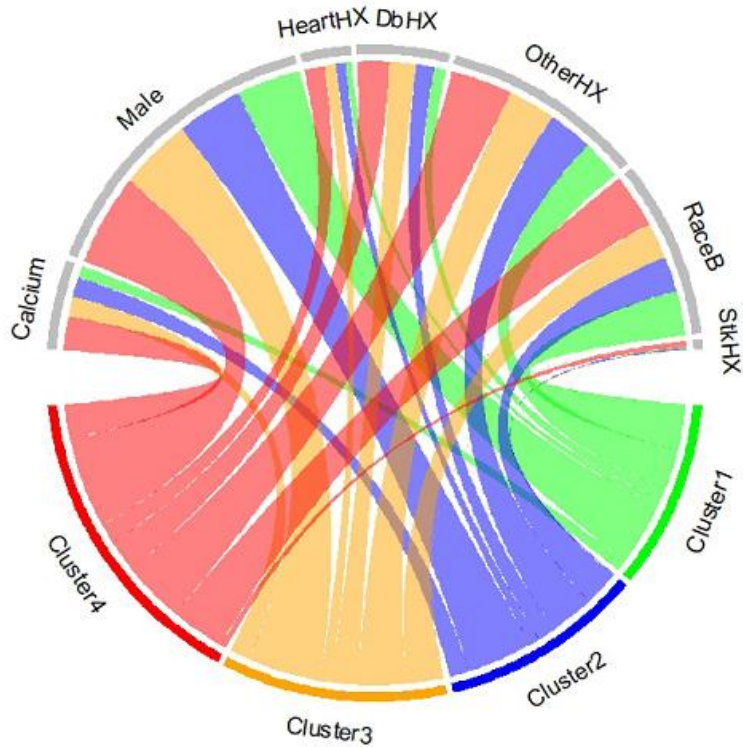

**Figure 3**. Chord diagrams showing clinical variables associated with different clusters. Note: Calcium-Calcium channel blockers, HeartHX: Heart disease history, DbHX: Diabetes history, OtherHX: Other disease history, RaceB: African American, StkHX: Stroke history.

**Discussion**

We identified subtypes of patients with OHT that experienced different courses of VF progression. In contrast to most of the previous studies, we first identified fast subtypes based on an unsupervised LCMM model to avoid subjectivity and bias in the analysis. We discovered four different subtypes of eyes (clusters 1 to 4) labeled as Improvers, Stables, Slow-progressors, and Fast-progressors, based on trajectories of their MD over about 11 years (average) follow up. From the VF trajectories presented in Figure 1, MD of the eyes in cluster 1 has improved slightly over time, thus this cluster was labeled as Improvers. The MD of the eyes in cluster 2 did not progress thus this cluster was labeled as Stables. The MD of the eyes in cluster 3 however deteriorated slowly thus we labeled this cluster as Slow-progressors. Finally, the MD of the eyes in cluster 4 deteriorated faster thus we labeled this cluster as Fast-progressors.

After identification of fast progressing and non-progressing subtypes, we investigated which baseline parameters may drive future fast VF progression. To the best of our knowledge, this is the first study to identify OHT subtypes based on unsupervised models investigating both linear and nonlinear trajectories of VF worsening.

Instead of using subjective thresholds to define progression (e.g., MD decline of -1 dB/year and worse)[22], we used unsupervised LCMM to objectively identify fast progressors without expert intervention. We then characterized subtypes and investigated factors that are associated with the fast progressor subtype. Identifying different OHT subtypes (with different courses of VF progression) is significant in real clinical settings since it can better inform clinicians to more frequently visit and monitor patients with higher risk of faster VF progression, tailor treatment plans, provide more accurate prognosis, and inform the development of new treatments and interventions [27,28]. Using LCMM for identifying subtypes is critical as progression is not always linear and may be nonlinear particularly at later stages of the disease [29], therefore a single MD rate may not fully represent progression, however, LCMM considers linear and nonlinear characteristics of the entire trajectory. LCMM has also been recently utilized to predict future VF loss in patients with glaucoma [22].

Based on subtype (cluster) characteristics and statistical analysis, we observed that subjects in the Fast-progressors cluster (subtype) were older, had thinner CCT, higher IOP, PSD and RE at baseline visit, compared to subjects in the Non-progressing subtype. As expected, age, IOP, PSD and RE were positively associated and CCT was negatively associated with fast VF progression. Moreover, a greater percentage of the patients in the Fast-progressors subtype eventually converted to glaucoma compared to subjects in the other subtypes, which further confirms the robustness of the LCMM in identifying fast progressors. Interestingly, we observed that patients with faster VF progression had a higher ratio of converting to glaucoma based on optic disc evaluations compared to VF abnormalities. Since optic disc abnormalities implies structure impairment and VF abnormalities implies functional impairment, our observation may imply structural defects drive fast VF progression [30]. However, structural and functional loss are correlated[31] and faster progression more likely leads to conversion to glaucoma. We also observed that some other clinic factors including calcium channel blockers, being male, heart disease history, diabetes history, stroke history, migraine headaches, and African American race were significantly different between Fast-progressors and non-progressors. The use of calcium channel blockers, being male, African American, having history of heart disease, diabetes, and stroke were positively associated with fast progression and migraine headaches was negatively associated with fast progression.

Some of these factors have been previously reported as risk factors of glaucoma progression. For instance, in agreement with our findings, age has been reported as a risk factor for glaucoma progression.[32] Likewise, IOP has been long reported as a risk factor of glaucoma and glaucoma progression.[4,32,33] Numerous studies have reported thinner CCT as a predictor of glaucoma progression [32-34] which agrees with our findings. While PSD[35] and RE [36,37] have been reported to be associated with glaucoma and its progression but the role of myopia in glaucoma progression is controversial (some studies suggest myopia as a risk factor for glaucoma progression and myopic eyes with higher than −4.0 D tended to progress faster[38]). While larger CDR serves as a hallmark of glaucoma[39], CDR and VCDR were similar among clusters. This may be explained by the fact that all OHTS participants had normal VF and optic disc appearance at the time of recruitment. Chan et al. [40] reported that glaucoma patients with fast progression had significantly higher rates of cardiovascular disease, which agrees with our findings.

Black race is long known as a risk factor for glaucoma and increased likelihood of progression.[41] We also observed that African American patients were more likely to experience future fast VF progression. Khachatryan et al. reported that males have higher likelihood of developing POAG.[42] We found that males may be at a higher risk of fast progression compared to females. In this study, we also identified some factors that were not previously reported as risk factors for glaucoma or glaucoma progression. These include taking calcium channel blockers and history of stroke.

One of the OHTS study goals was to evaluate the safety and efficacy of topical ocular hypertensive medication in delaying or preventing the development of POAG in patients with OHT. The OHTS has shown that topical medications can reduce IOP about 20% and can prevent or delay progression significantly [23]. The percentage of eyes that received medication in clusters Improvers, Stables, Slow-progressors, and Fast-progressors were 51.3%, 48.3%, 51.8% and 42.9%, respectively, which was not statistically different among four clusters (Table 5). It is worth mentioning that all OHTS participants took medicine from the second phase (after June 2002) of the study. Other studies may investigate the impact of medication on the course of glaucoma progression.

This study can assist physicians adjust treatment plans in multiple ways. For example:1) Patients with higher IOP, higher age, higher PSD and RE, lower CCT, taking calcium channel blockers, male, heart disease history, diabetes history, African American race, or stroke history have a higher chance to progress faster thus closer management or treatment regimen are required. 2) Use the information of our four subtypes (mean of different parameters in these subtypes we provided in Tables) and identify the factors of a OHT patient is closer to which subtype then assess the risk of future faster progression. 3) Also take into account the impact of two novel risk factors for faster VF progression.

Our study has numerous limitations as well. First, the study participants were selected from the OHTS in which only patients with OHT participated. As such, the findings may not be generalizable to other POAG patients or the whole POAG population. Second, as the subjects had normal VF and normal appearing optic disc at the bassline, a relatively small percentage of the eyes progressed to glaucoma. Third, the course of glaucoma was limited to mostly early stages and progression to moderate or advanced stage of POAG were rare. Follow-up studies with more diverse POAG cases along a wider spectrum of the disease severity are warranted to support findings. Lastly, we selected reliable VFs for the downstream analysis, but VF variability might also influence VF trend. Investigating the influence of VF variability can be performed in follow up studies.

**Conclusion**

We discovered various subtypes of OHT patients with different VF progression based on an unsupervised clustering model without any expert intervention or utilization of subjective thresholds for identifying progression. Based on this unbiased and objective analysis, we observed that older age, higher IOP and PSD and lower CCT at the baseline were associated with future fast VF progression. Additionally, taking calcium channel blockers, being male, being African American, having history of heart disease, diabetes, and stroke were also associated with future fast VF progression. This study can lead to automated ways for selection of appropriate candidates for glaucoma clinical trials and provide guidance for early treatment of glaucoma patients to delay vision loss and maintain vision-related quality of life.

**Figures and Tables**

**Figure 1**. Four clusters (subtypes) identified by the LCMM model. Smoothed trajectory of MD of the eyes in four clusters representing Improvers (green), Stables (blue), Slow progressors (brown), and Fast progressors (red).

**Figure 2.** Survival analysis of the eyes in different clusters.

**Figure 3**. Chord diagrams showing clinical variables associated with different clusters. Note: Calcium-Calcium channel blockers, HeartHX: Heart disease history, DbHX: Diabetes history, OtherHX: Other disease history, RaceB: African American, StkHX: Stroke history.

**Table 1.** Demographic characteristics of the participants (eyes) at the baseline visit. Continuous variables are summarized based on mean and standard deviation (SD) while categorical variables are presented as counts and frequencies (%).

**Table 2.** Clustering membership evaluations based on different subsets of eyes and different number of visual fields.

**Table 3.** Intercept and slope of the model for each cluster (subtype).

**Table 4.** Characteristics of the baseline parameters of the eyes in each cluster (continuous variables) presented as mean (SD). IOP: Intraocular pressure; CCT: Central corneal thickness; MD: Mean deviation; PSD: Pattern standard deviation; CDR: Cup disk ratio; VCDR: Vertical cup disc ratio; RE: Refractive error. a: significance among all 4 clusters; b: difference between cluster 4 and combined clusters 1 and 2; c: odds ratio of cluster 4.

**Table 5.** Characteristics of the baseline parameters of eyes in different clusters (categorical variables).
**a**: difference among all 4 clusters; **b**: difference between cluster 4 and combined clusters 1 and 2; **c**: odds ratio of cluster 4. POAG: Primary open-angle glaucoma; VF: Visual field.

**Acknowledgment**

This work was supported by NIH Grants R01EY033005 (SY), R21EY031725 (SY), and Challenge Grant from Research to Prevent Blindness (RPB), New York (SY). The funders had no role in study design, data collection and analysis, decision to publish, or preparation of the manuscript.